\documentclass[10pt,twocolumn,letterpaper]{article}
\usepackage{multirow}

\usepackage[pagenumbers]{cvpr}

\definecolor{cvprblue}{rgb}{0.21,0.49,0.74}
\usepackage[pagebackref,breaklinks,colorlinks,allcolors=cvprblue]{hyperref}

\usepackage{algorithm}
\usepackage{algpseudocode}

\title{FluxSpace: Disentangled Semantic Editing in Rectified Flow Transformers}

\author{Yusuf Dalva \qquad
Kavana Venkatesh \qquad
Pinar Yanardag \\
Virginia Tech \\
\small{\url{https://fluxspace.github.io}}
}

\begin{document}

\twocolumn[{
\maketitle
\begin{center}
    \captionsetup{type=figure}
    \vspace{-1em}
\newcommand{\imwidth}{1\textwidth}
\begin{tabular}{@{}c@{}}

\parbox{\imwidth}{ \centering \includegraphics[width=0.9\imwidth, ]{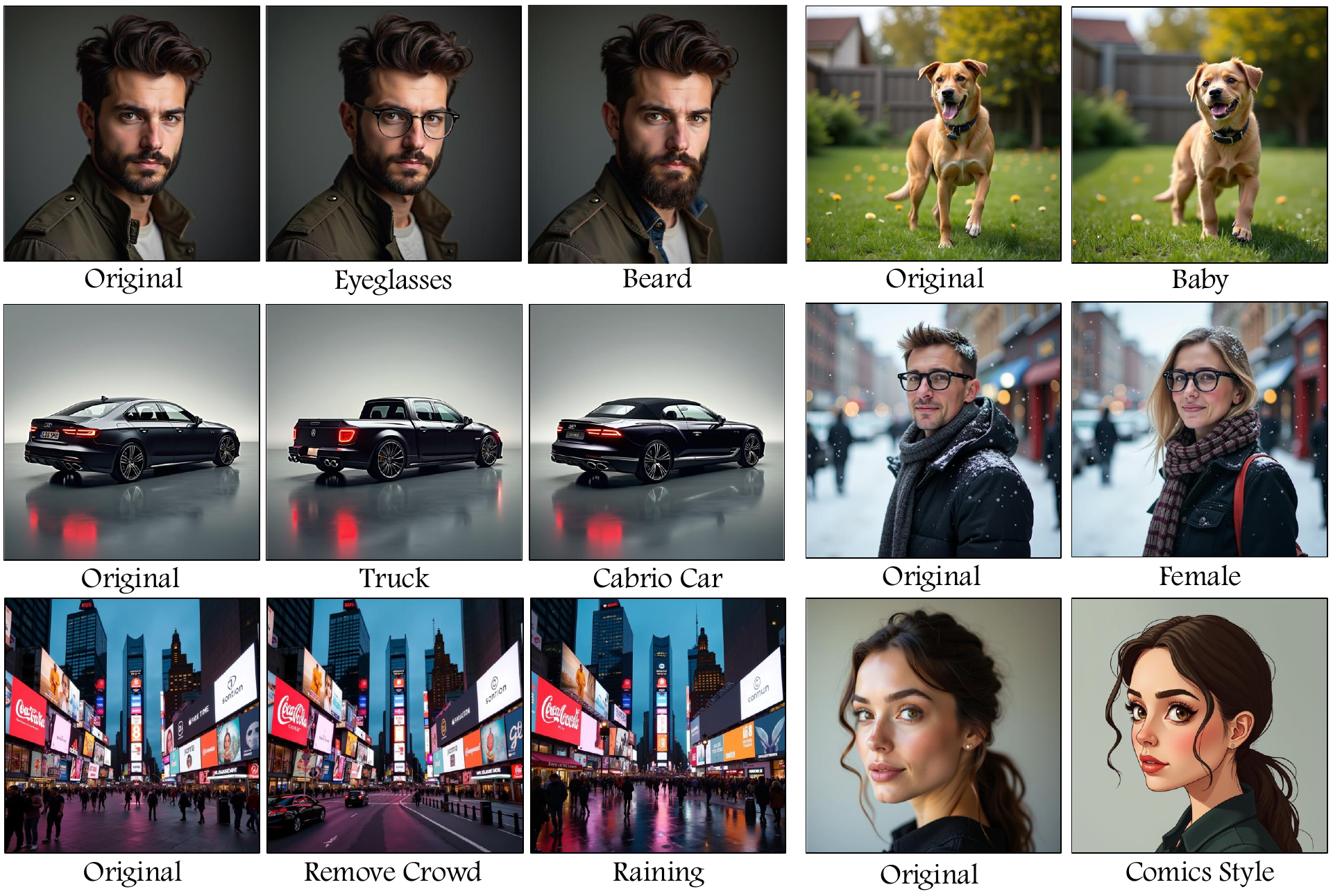}}
\\

\vspace{1em}
\end{tabular}
    \vspace{-2.5em}
    \captionof{figure}{\textbf{FluxSpace.} We propose a text-guided image editing approach on rectified flow transformers \cite{esser2024scaling}, such as Flux. Our method can generalize to semantic edits on different domains such as humans, animals, cars, and extends to even more complex scenes such as an image of a street (third row, first example). FluxSpace can apply edits described as keywords (e.g. ``truck'' for transforming a car into a truck) and offers disentangled editing capabilities that do not require manually provided masks to target a specific aspect in the original image. In addition, our method does not require any training and can apply the desired edit during inference time.} 
    \label{fig:teaser}
\end{center}
}]

\maketitle
\begin{abstract} 
Rectified flow models have emerged as a dominant approach in image generation, showcasing impressive capabilities in high-quality image synthesis. However, despite their effectiveness in visual generation, rectified flow models often struggle with disentangled editing of images. This limitation prevents the ability to perform precise, attribute-specific modifications without affecting unrelated aspects of the image. In this paper, we introduce FluxSpace, a domain-agnostic image editing method leveraging a representation space with the ability to control the semantics of images generated by rectified flow transformers, such as Flux. By leveraging the representations learned by the transformer blocks within the rectified flow models, we propose a set of semantically interpretable representations that enable a wide range of image editing tasks, from fine-grained image editing to artistic creation. This work offers a scalable and effective image editing approach, along with its disentanglement capabilities.
\end{abstract}    
\section{Introduction}
\label{sec:intro}

Recent research aimed at enhancing the interpretability of generative models has increasingly focused on disentangled editing capabilities, which allows precise control over specific features or attributes within generated images \cite{wu2021stylespace, yuksel2021latentclr, stylegan}. These capabilities are crucial for understanding and manipulating the outputs of generative models, such as Generative Adversarial Networks (GANs) and multi-step diffusion models like Stable Diffusion. In single-step models such as GANs \cite{goodfellow2020generative}, the editing process is facilitated by a structured and highly disentangled latent space. This space, composed of fixed-dimension vectors, linearly encodes meaningful concepts that can be individually adjusted to effectively alter specific aspects of the output images. However, achieving similar disentangled editing in multi-step diffusion models presents a significant challenge. Unlike GANs, diffusion models do not utilize a fixed-dimension latent space; instead, they generate images through a multi-step refinement process. This process gradually transitions from a random noise distribution to the data distribution, with each step involving complex interactions of learned noise patterns. These patterns are inherently difficult to map back to specific features in a latent space.  Moreover, the sequential nature of diffusion models complicates the internal dynamics and obscures the direct correlations between dimensions of a hypothetical latent space and specific visual features in the generated output.  

Several studies have explored disentangled latent spaces in diffusion models, focusing on the UNet bottleneck layer \cite{kwon2022diffusion, park2023understanding}, text embedding space \cite{baumann2024continuous}, noise space \cite{dalva2023noiseclr} and weight space \cite{dravid2024interpreting, gandikota2023concept}.  However, these approaches often suffer from significant limitations in terms of disentangled semantics. Manipulation of the UNet bottleneck layer \cite{kwon2022diffusion, park2023understanding} may not fully capture the high-level semantic features necessary for fine-grained control, as bottleneck representations can still be highly entangled. Adjusting the text embedding space \cite{baumann2024continuous} relies heavily on alignment between textual and visual features, which may not always correspond accurately, leading to entangled semantics. Operating within the noise space \cite{dalva2023noiseclr} presents challenges, as it is difficult to associate specific noise patterns with distinct semantic attributes. Other works \cite{dravid2024interpreting} explored the weight space to perform disentangled editing; however, they require training of LoRA models to create a latent space for each domain, which is time and resource consuming. On the other hand, generative models based on flow-matching transformers (e.g. Flux) offer high-quality image generation; however, their disentanglement editing capabilities remain underexplored.

In this paper, we focus on the latent spaces in flow-matching transformers, characterized by their ability to generate images with high fidelity \cite{esser2024scaling}. Our analysis reveals that the joint transformer blocks, integral to the denoising network, are adept at encoding highly disentangled semantic information. Since the architecture we target involves transformers, the attention layer outputs gradually add image content to the latent representation that gets denoised, without any residual connections between different blocks, unlike UNet-based architectures. Furthermore, the attention layers included in these blocks enable enhanced control over the model outputs, as they control the image content in isolation compared to succeeding and preceding blocks of the denoising network. By implementing a linear editing scheme across these attention outputs, we unlock the potential for semantic editing within flow-matching transformers, facilitating semantic navigation across their output space. Our approach supports precise, fine-grained edits, such as adding a smile, as well as broader, coarse-level modifications, such as stylization.  We perform qualitative and quantitative experiments with various state-of-the-art methods and demonstrate the effectiveness of our method in enabling fine-grained image editing tasks. Our contributions are as follows.

\begin{itemize}
    \item We introduce FluxSpace, a novel editing framework within flow-matching transformers using attention layer outputs, unlocking advanced semantic editing capabilities for precise navigation within the model's output space.
    \item We demonstrate the capability of joint transformer blocks to encode highly disentangled semantic information through incremental content refinement.
    \item We introduce support for both fine-grained edits, such as smile addition, and coarse-level modifications such as stylization, improving the model's versatility in image manipulation. Our method supports the editing of both real and generated images.
    \item We make our implementation public to facilitate research in this area.
\end{itemize}

\section{Related Work}

\paragraph{Latent Space Exploration of Diffusion Models.}
Diffusion models, which have been the dominant approach for the tasks of text-to-image generation and editing, encapsulate rich semantics within their latent representations. To facilitate new applications, research has focused on leveraging the semantics encoded by such models. In addition to the explorations performed on latent spaces, certain methods \cite{kwon2022diffusion, wu2023latent} explored image editing techniques that alter the backward diffusion process, targeting the learned latent variables. Specifically, \cite{kwon2022diffusion} bases its approach on the features learned by the bottleneck block of the denoising model, which has a UNet-based architecture, while \cite{wu2023latent} modifies the intermediate latent variables using stochastic diffusion models. As a further advancement toward this area, \cite{park2023understanding} introduces a method that aims to identify latent-specific directions representing various semantics, inspired by latent space exploration methods in GANs \cite{yuksel2021latentclr, kocasari2022stylemc, dalva2023image, patashnik2021styleclip}. Although such methods promise the discovery of semantic directions in domain-specific DDPMs \cite{ho2020denoising}, their application to large-scale diffusion models \cite{rombach2022high, podell2023sdxl, esser2024scaling} is limited. In addition, various approaches have studied alternative latent spaces to propose directions for semantic editing, such as the noise space \cite{dalva2023noiseclr}, the text embedding space \cite{baumann2024continuous}, and the weight space \cite{gandikota2023concept, dravid2024interpreting}. Despite these efforts that propose different formulations of semantic directions on diffusion models, such an approach still does not exist for flow-matching transformer models, which is addressed by our proposed FluxSpace framework.

\paragraph{Image Editing with Diffusion Models.}
Due to their high-fidelity generation capabilities, text-to-image diffusion models have been used frequently for image generation and editing tasks. As a straightforward strategy, one can perform image editing with a target text prompt that effectively describes the desired visual effect. However, this strategy can lead to entangled edits, where the change described as a text prompt modifies certain aspects of the image that are not intended to be edited. Addressing this problem, studies such as \cite{hertz2022prompt, zhang2023adding} enabled increased precision within the editing process. Among these works, \cite{zhang2023adding} uses a conditional diffusion model that guides the generation process based on user-specified controls. Similarly, \cite{valevski2023unitune} attempts to preserve image content by fine-tuning the diffusion model. On the other hand, several methods benefit from the representations learned by the diffusion models. Among these methods, \cite{hertz2022prompt, tumanyan2023plug} use attention control mechanisms within the diffusion model, \cite{brack2023sega, brack2024ledits++} use the semantic guidance mechanism over the noise space, and \cite{baumann2024continuous, yesiltepe2024curious} utilize the text-embedding representations to perform edits. Furthermore, studies such as \cite{dalva2024gantastic, gandikota2023concept} succeeded in applying disentangled edits by trained semantic directions. However, these methods are limited in terms of requiring training-per-edit to perform the desired image-to-image transformation, which is either specified as a set of textual conditions or paired images reflecting the desired edit. Addressing the problems in both directions, we focus on disentangled editing on rectified flow transformers, without requiring any additional training.

\paragraph{Rectified Flow-Based Models.}
Succeeding over diffusion models, transformers trained with the flow-matching objective \cite{lipman2023flow, liu2023flow}, such as Flux, now serve as the state-of-the-art text-to-image generation models \cite{esser2024scaling}. Despite their superior generation capabilities, the methods applied to diffusion models for image editing are not directly transferable to rectified flow models, since the MM-DiT architecture \cite{peebles2023scalable} involves continuous interaction between text and image features, which results in changes in both representations, unlike diffusion models \cite{rombach2022high, podell2023sdxl}. Furthermore, disentangled editing with existing methods is not possible in a straightforward manner. As a preliminary effort towards text-based image editing with rectified flows, \cite{rout2024rfinversion} offers an editing method with target captions specifying the desired edit. However, existing methods are limited in terms of the types of edits available and edit scale adjustment, whereas our method offers such functionalities.

\section{Preliminaries}

\begin{figure*}[h!]
    \centering
    \includegraphics[width=\linewidth]{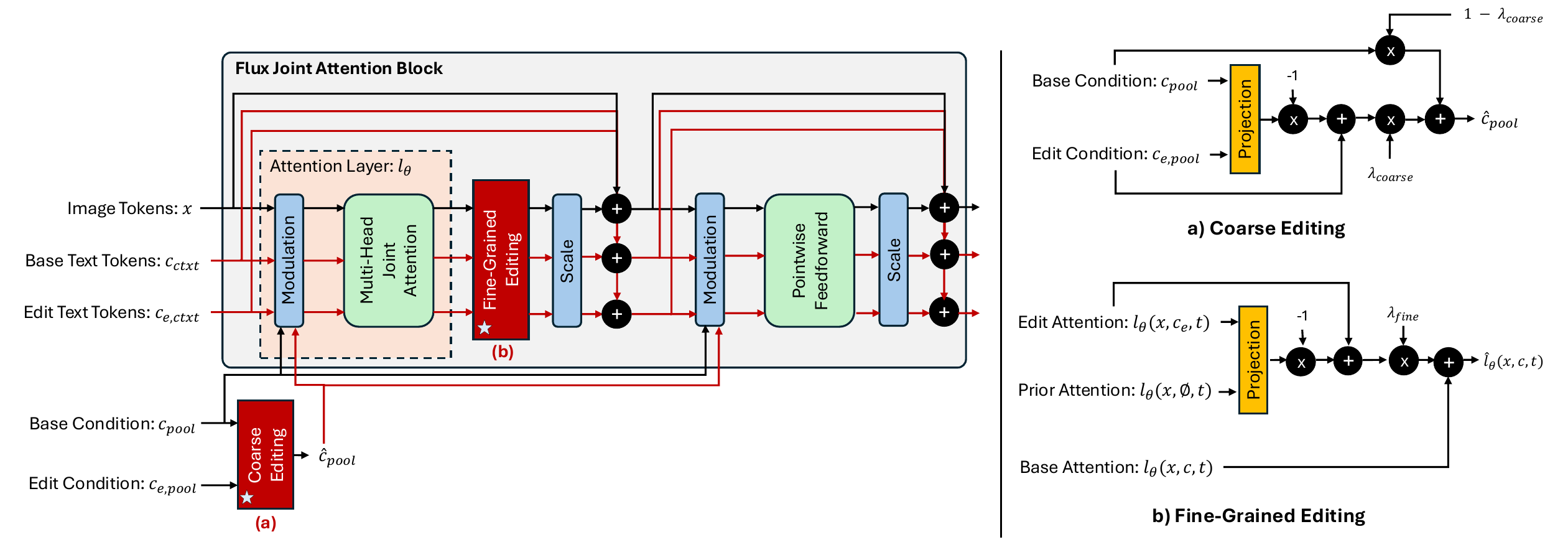}
    \caption{\textbf{FluxSpace Framework.} The FluxSpace framework introduces a dual-level editing scheme within the joint transformer blocks of Flux, enabling coarse and fine-grained visual editing. Coarse editing operates on pooled representations of base ($c_{pool}$) and edit ($c_{e, pool}$) conditions, allowing global changes like stylization, controlled by the scale $\lambda_{coarse}$ (a). For fine-grained editing, we define a linear editing scheme using base, prior, and edit attention outputs, guided by scale $\lambda_{fine}$ (b). With this flexible design, our framework is both able to perform coarse-level and fine-grained editing, with a linearly adjustable scale.}
    \label{fig:framework}
\end{figure*}

\subsection{Rectified-Flow Models}
Generative models aim to define a mapping from samples \( x_1 \) from a noise distribution \( p_1 \) to samples \( x_0 \) from a data distribution \( p_0 \), where \( p_0 \) represents real images in image generation tasks. Rectified flows \cite{lipman2023flow, liu2023flow} define a forward process that constructs paths between distributions \( p_0 \) and \( p_1 \) as straight trajectories, as shown in Eq. \ref{eqn:flow_path}, where \( p_1 = \mathcal{N}(0, 1) \). Here, the forward process is time-dependent due to timestep \( t \).

\begin{equation}
    \label{eqn:flow_path}
    x_t = (1 - t)x_0 + t\epsilon, \quad \epsilon \sim N(0, 1)
\end{equation}

\noindent To learn this mapping, a network is trained with parameters $\theta$, to estimate the velocity \( v \) of the rectified flow, represented by \( v_\theta \). By adopting the reparameterization from \cite{esser2024scaling}, this velocity prediction network can serve as a noise prediction network, \( \epsilon_\theta \), optimized using the Conditional Flow Matching (CFM) objective formulated as Eq. \ref{eqn:cfm_objective}.

\begin{equation}
    \label{eqn:cfm_objective}
    \mathcal{L}_{CFM} = -\frac{1}{2} \mathbb{E}_{t \sim \mathcal{U}(t), \epsilon \sim \mathcal{N}(0, I)}[w_t \lambda_t'||\epsilon_{\theta}(x_t, t) - \epsilon||^2]
\end{equation}

\noindent Here, \( \lambda_t' \) represents the re-parametrized signal-to-noise ratio, and \( w_t \) is a time-dependent weighting function.  

\subsection{Multi-Modal Diffusion Transformers}

The multi-modal diffusion transformer architecture, multi-modal diffusion transformers (MM-DiT) \cite{peebles2023scalable}, integrates both text and image modalities to generate images aligned with the semantics implied by text inputs. Among the state-of-the-art text-to-image generation models, rectified flow transformers (e.g. Flux) utilize the adaptation of this architecture introduced in \cite{esser2024scaling}, which involves modulated attention and MLP layers, followed by attention layers $l_\theta$ parametrized by parameters $\theta$, conditioning image generation on both pooled and token-wise text embeddings.

To guide image generation with text inputs, rectified flow transformers use two text embeddings: \( c_{pool} \) and \( c_{ctxt} \). The pooled embedding \( c_{pool} \), derived from the CLIP Text Encoder \cite{radford2021learning}, is used in the modulation mechanism to scale and shift features fed as input to the attention layers. The token-wise embeddings \( c_{ctxt} \), obtained from a T5 Text Encoder \cite{raffel2020exploring}, ensure alignment with prompt semantics, thus enhancing the relevance of the generated images. The complete conditioning set, \( \{c_{pool}, c_{ctxt}\} \), improves prompt alignment, as shown in previous work \cite{podell2023sdxl, esser2024scaling}. Both embeddings are integrated into joint transformer blocks, where text and image features interact through distinct query ($Q$), key ($K$), and value ($V$) transformations. These blocks enable influence across modalities (text and image) in a bidirectional manner, forming the foundation for the image editing mechanism introduced in this work.

\section{Methodology}
\label{sec:methodology}

\begin{figure*}[h!]
    \centering
    \includegraphics[width=0.97\linewidth]{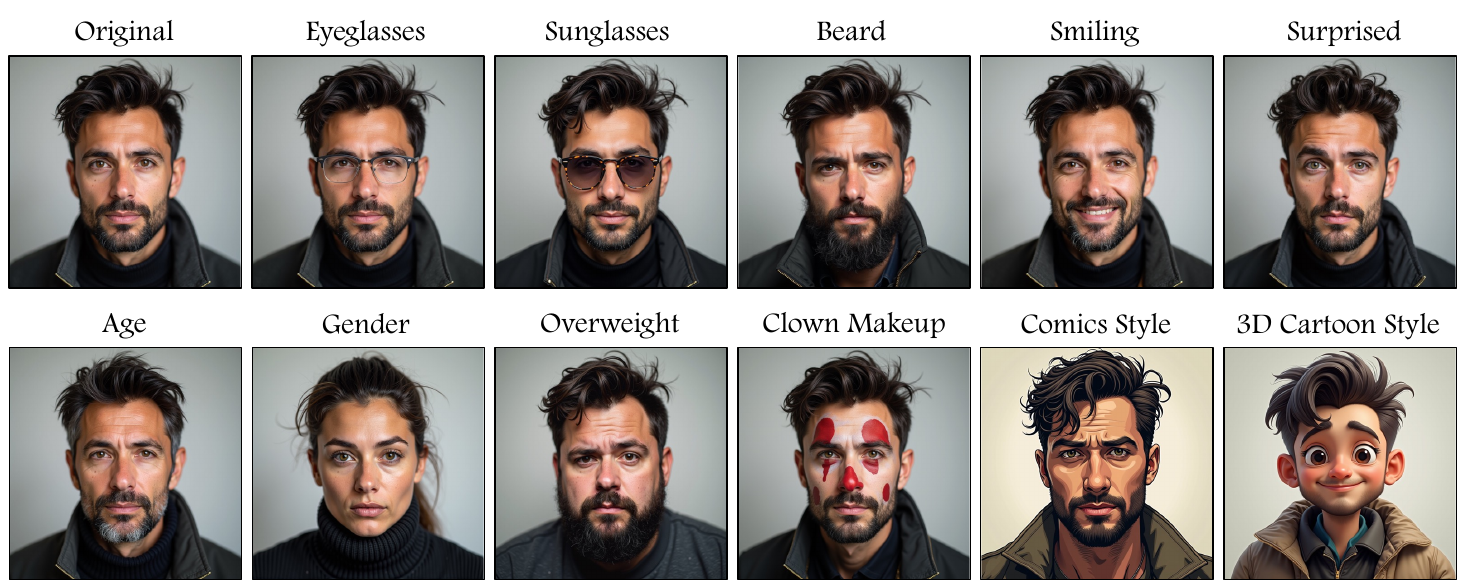}
    \caption{\textbf{Qualitative Results on Face Editing.} Our method can perform a variety of edits from fine-grained face editing (e.g. adding eyeglasses) to changes over the overall structure of the image (e.g. comics style). As our method utilizes disentangled representations to perform image editing, we can precisely edit a variety of attributes while preserving the properties of the original image.}
    \label{fig:face_edits}
\end{figure*}

Our approach leverages the architecture of flow-matching transformers, specifically Flux, to introduce a structured method for controlled image editing. In Flux, image generation progresses by refining content from initial noise through multi-modal transformer blocks. At each timestep \( t \), the noise prediction network \( \epsilon_\theta(x_t, c, t) \) outputs a prediction conditioned on noisy latent \( x_t \) and text-based conditioning \( c \) (composed of \( c_{pool} \) and \( c_{ctxt} \)). This stepwise refinement occurs within the attention layers \( l_\theta \), creating a progressively evolving representation space suitable for semantic manipulations. For simplicity, we represent the conditional input to the attention layer as $c$, where it contains both the modulating text embedding $c_{pool}$ and the token-wise text condition $c_{ctxt}$, which both influences the attention mechanism within Flux.

Our method designates the outputs of Flux’s joint attention layers as \textit{FluxSpace}, a linear representation space where semantic image edits can be performed in a disentangled manner. Within this space, we enable transformations on the attention outputs, allowing semantic modifications such as detailed adjustments to object attributes or overall changes in style, without altering Flux’s pretrained parameters.

Our primary objective is to introduce an inference-time image editing algorithm with flexible levels of control. The first option focuses on detailed edits, allowing fine-grained adjustments, such as adding a smile to a face. The second option provides control over the image's coarse appearance, such as changing the overall style.

\subsection{Fine-Grained Editing}

Given the pre-trained attention layer $l_\theta$ and the image tokens from the noisy image input $x_t$, we employ three different outputs obtained by this layer. With a minor use of notation, we use $x$ for image tokens from the noisy input, both for simplicity and to illustrate the relation between $x_t$ and image tokens. First, we utilize the output $l_\theta(x, c, t)$ where $c$ is the text condition used to generate the unedited image. Following, we compute two additional features, $l_\theta(x, c_e, t)$ and $l_\theta(x, \phi, t)$ corresponding to the outputs w.r.t. editing condition $c_e$ (e.g. ``eyeglasses'' for the addition of eyeglasses) and null text $\phi$. In our editing scheme, we treat $l_\theta(x, \phi, t)$ as an image prior, which reflects the attention layers' knowledge of the image features without any supplementary text conditions.

Our framework relies on the linearity assumption of attention outputs, which enables us to define latent directions on a given input condition $c_e$. First, to isolate the conditional output $l_\theta(x_t, c_e, t)$ from the image-related details, we apply an orthogonal projection with the image prior retrieved as the null prediction $l_\theta(x_t, \phi, t)$, to obtain $\text{proj}_{\phi}l_\theta(x, c_{e}, t)$ formulated over Eq. \ref{eqn:projection}.

\begin{equation}
    \label{eqn:projection}
    \text{proj}_{\phi}l_\theta(x, c_{e}, t) = \frac{l_\theta(x, c_{e}, t) \cdot l_\theta(x, \phi, t)}{||l_\theta(x, \phi, t)||^2} l_\theta(x, \phi, t)
\end{equation}

Given the projection of attention outputs, we identify the orthogonal component of $l_\theta(x, c_e, t)$ over $l_\theta(x, \phi, t)$ as $l'_\theta(x, c_e, t)$ in Eq. \ref{eqn:ortho}. With this vector, we identify a semantic direction that is effective in latent pixels to shift image content w.r.t. editing prompt $c_e$.

\begin{equation}
    \label{eqn:ortho}
    l'_\theta(x, c_{e}, t) = l_\theta(x, c_{e}, t) - \text{proj}_{\phi}l_\theta(x, c_{e}, t)
\end{equation}

Using this linear direction, we formulate our editing scheme in Eq. \ref{eqn:edit_content}. As we define our editing method in the form of linear interpolation between the editing vector $l'_\theta(x, c_e, t)$, our method is also able to perform edit interpolation with the editing scale $\lambda_{fine}$, which controls the strength of the edit.

\begin{equation}
    \label{eqn:edit_content}
    \hat{l_\theta}(x, c, c_{e}, t) = l_\theta(x, c, t) + \lambda_{fine} (l'_\theta(x, c_{e}, t))
\end{equation}

\noindent \textbf{Content Preservation with Attention Masking.}
To facilitate further disentanglement over the performed edit, we introduce a self-supervised mask, based on the interaction of the image features and the editing condition. First, we introduce the mask $M_{i, edit}$ based on the query features $Q_i$, computed with the image features, and the key features $K_{edit}$ with the editing condition in Eq. \ref{eqn:attn_mask}. Intuitively, we query the image for the pixels that respond strongly to the given text condition as an intermediate estimate, which is used to mask out the pixels with low attention values. Following the existing work on the T5 text encoder, which is used in Flux, we utilize the attention map of the first text token \cite{ni2021sentence}.

\begin{equation}
    \label{eqn:attn_mask}
    M_{i, edit} = softmax\Bigg(\frac{Q_i \cdot K_{edit}^T}{\sqrt{d}}\Bigg)
\end{equation}

Given the mask $M_{i, edit}$, we introduce a soft decision boundary $M_{i, edit}'$ with boundary coefficient $d = 10$ \cite{epstein2023selfguidance}, sigmoid operator $\sigma$, and min-max normalization operator $normalize$, formulated as Eq. \ref{eqn:decision_boundary}. 

\begin{equation}
    \label{eqn:decision_boundary}
    M_{i, edit}' = \sigma(d * (normalize(M_{i, edit}) - 0.5))
\end{equation}

As the final step, we perform a thresholding operation with the parameter $\tau_{m}$ to retrieve the thresholding mask $M_{i, edit}''$ in Eq. \ref{eqn:threshold}.

\begin{equation}
    \label{eqn:threshold}
    M_{i, edit}'' = 
\begin{cases} 
   1 & \text{if } M_{i, edit}' \geq \tau_{m} \\
   0 & \text{otherwise} 
\end{cases}
\end{equation}

Using this thresholding mask, we modify the content editing equation presented in Eq. \ref{eqn:edit_content} by masking out the latent pixels that receive low attention from the editing direction $l'_\theta(x, c_{e}, t)$.

\begin{figure*}[t!]
    \centering
    \includegraphics[width=0.95\linewidth]{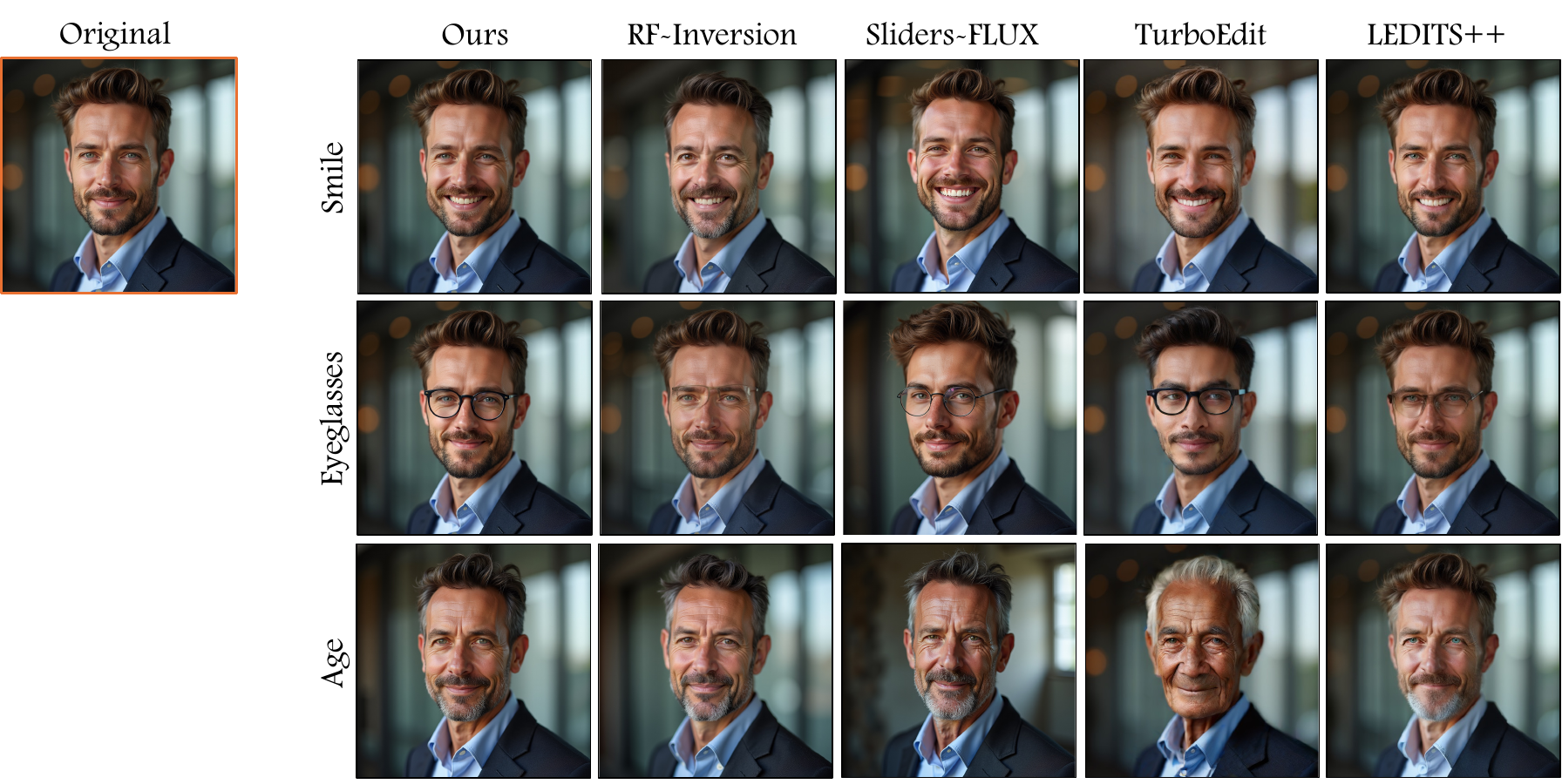}
    \caption{\textbf{Qualitative Comparisons.} We compare our method both with latent diffusion-based approaches (LEDITS++ \cite{brack2024ledits++} and TurboEdit \cite{deutch2024turboedittextbasedimageediting}) and flow-based methods (Sliders-FLUX \cite{gandikota2023concept} and RF-Inversion \cite{rout2024rfinversion}) in terms of their disentangled editing capabilities. We present qualitative results for smile, eyeglasses, and age edits where our method succeeds over competing methods in both reflecting the semantic and preserving the input identity.}
    \label{fig:qualitative}
\end{figure*}

\subsection{Editing Coarse Level Details}
\label{sec:coarse_edit}
Before performing the attention calculation, Flux applies a modulation based on the pooled CLIP embeddings that provides information about the coarse structure of the generated image \cite{podell2023sdxl}. Since certain edits need to change the overall structure and appearance of the image, we introduce an additional control mechanism for appearance-based changes. Specifically, we extend our editing approach based on the orthogonal projection of attention features in the CLIP embeddings, where we edit the pooled generation condition $c_{pool}$ in the direction of the pooled editing condition $c_{e, pool}$ to retrieve $\hat{c}_{pool}$, which is used to normalize the attention features.

Since we want to edit the coarse condition in a disentangled manner, we perform our editing scheme based on the application of the linear representation hypothesis \cite{mikolov2013linguistic, park2023the, arora2018linear, arora2016latent, faruqui2015sparse} on pooled CLIP embeddings \cite{bhalla2024interpreting}. To do so, we first perform an orthogonal projection in Eq. \ref{eqn:pooled_emb_proj}, for the editing concept $c_{e, pool}$ on $c_{pool}$ to retrieve the editing direction for the coarse representation of the image. However, different from the attention features, we use the base generation condition as the basis of the projection, as we now operate on textual representations rather than multimodal representations. 

\begin{equation}
    \label{eqn:pooled_emb_proj}
    c'_{e, pool} = c_{e, pool} - \frac{c_{pool} \cdot c_{e, pool}}{||c_{pool}||^2} c_{pool}
\end{equation}

Given the linear direction for editing the coarse condition, we formulate the edit as a linear interpolation between the original condition and the editing direction with a scale parameter $\lambda_{coarse}$, which controls the extent of editing of the modulation condition.

\begin{equation}
    \label{eqn:pooled_emb_edit}
    \hat{c}_{pool} = (1 - \lambda_{coarse}) * c_{pool} + \lambda_{coarse} * c'_{e, pool}
\end{equation}

During generation, to preserve the image content and influence the image with the desired edit, we use different modulations for text and image features. Specifically, we normalize the image features inputted to the attention layer $l_\theta$ with $c_{pool}$, while we modulate the text features with $\hat{c}_{pool}$, which enables changes in the overall appearance of the image.

\section{Experiments}
\definecolor{lightorange}{rgb}{1.0, 0.9, 0.7}

To assess the effectiveness of FluxSpace in disentangled semantic editing, we conducted a series of qualitative and quantitative experiments. Furthermore, we compared our approach with the state-of-the-art image editing methods in both flow-based and latent diffusion-based models to highlight our method's versatility and superior performance.

\begin{table*}[h!]
\centering
\begin{tabular}{l|lll|lll|c}
\toprule
\multirow{2}{*}{} & \multicolumn{3}{c|}{Eyeglasses} & \multicolumn{3}{c|}{Smile} & Overall\\ \hline
                  & \multicolumn{1}{c}{CLIP-T $\uparrow$} & \multicolumn{1}{c}{CLIP-I $\uparrow$} & \multicolumn{1}{c|}{DINO $\uparrow$} & \multicolumn{1}{c}{CLIP-T $\uparrow$} & \multicolumn{1}{c}{CLIP-I $\uparrow$} & \multicolumn{1}{c|}{DINO $\uparrow$} & User Pref. $\uparrow$ \\ \hline
LEDITS++ \cite{brack2024ledits++} & \textbf{0.3723} & 0.8639 & 0.9177 & 0.3634 & 0.8657 & 0.9111 & 3.04 \\
TurboEdit \cite{deutch2024turboedittextbasedimageediting} & 0.3626 & 0.8315 & 0.8954 & \textbf{0.3692} & 0.8479 & 0.9082 & 2.75\\
Sliders-FLUX \cite{gandikota2023concept} & 0.2878 & 0.8356 & 0.7858 & 0.3618 & 0.8268 & 0.8683 & 3.34 \\
RF-Inversion \cite{rout2024rfinversion} & 0.3046 & 0.9354 & 0.9359 & 0.3438 & 0.8834 & 0.9134 & 2.91 \\ 
\hline
\textbf{Ours} & 0.3180 & \textbf{0.9417} & \textbf{0.9402} & 0.3503 & \textbf{0.9038} & \textbf{0.9347} & \textbf{4.19} \\
\bottomrule
\end{tabular}
\caption{\textbf{Quantitative Results.} We quantitatively measure the editing performance of our method over competing approaches both in terms of text alignment using CLIP-T \cite{radford2021learning}, and content preservation using CLIP-I \cite{radford2021learning} and DINO \cite{caron2021emerging} metrics where higher is better for all metrics. We compare our method with both latent diffusion \cite{brack2024ledits++, deutch2024turboedittextbasedimageediting}, and flow-matching-based approaches \cite{gandikota2023concept, rout2024rfinversion}. Overall, our method strikes a good balance in terms of alignment with the editing prompt and content preservation. Supplementary to these metrics, we also present a user study as a perceptual evaluation that aligns with our claims regarding edit performance, where our method outperforms the competing approaches.}
\label{tab:quantitative}
\end{table*}

\subsection{Experimental Setup}
We utilize \texttt{FLUX.1-dev}\footnote{\url{https://huggingface.co/black-forest-labs/FLUX.1-dev}} for all experiments, evaluating both the capabilities of the proposed framework and the control parameters introduced. For all of our experiments, we use 30 generation steps and $\lambda_{coarse} = 0.5$ and $\tau_{m} = 0.5$ unless stated otherwise. Since the required editing strength changes for every editing task, we use varying values of $\lambda_{fine}$ and starting timestep for every experiment. We provide the complete list of used hyperparameters in the supplementary material. To guarantee the reproducibility of our experiments, we maintained a consistent random seed of 0. Our method requires approximately 20 seconds to generate an image with the desired edit on a single NVIDIA L40 GPU.

\subsection{Qualitative Results} Fig. \ref{fig:face_edits} and \ref{fig:qualitative} show the qualitative results of our editing experiments. These results demonstrate our method's capability to execute disentangled edits on fine-grained attributes, such as adding eyeglasses or sunglasses, as well as beards, smiles, and detailed facial expressions such as surprise (see Fig. \ref{fig:face_edits}, top row). Additionally, our method can conduct broader transformations, including altering a person's perceived age or gender and applying stylistic changes like converting images to comic or 3D cartoon styles (see Fig. \ref{fig:face_edits}, bottom row). Beyond faces, our method handles broader image contexts, where we provide examples in the supplementary material. Our approach also extends to other domains, including cars and complex scenes such as street views (see Fig. \ref{fig:teaser}). 

Furthermore, we conducted a qualitative comparison with several state-of-the-art methods, including approaches based on latent diffusion models like  LEDITS++ \cite{brack2024ledits++} (with SDXL \cite{podell2023sdxl}), and TurboEdit \cite{deutch2024turboedittextbasedimageediting} (with SDXL-Turbo \cite{sauer2025adversarial}), as well as flow-based methods such as Sliders-FLUX \cite{gandikota2023concept} and RF-Inversion \cite{rout2024rfinversion}. Fig. \ref{fig:qualitative} illustrates that our method consistently achieves disentangled edits, such as adding eyeglasses or smiles without altering unrelated features, while other methods often struggle with this aspect. For instance, RF-Inversion incorrectly positions the eyeglasses, placing them in unnatural locations on the face. Similarly, while performing edits, Concept Sliders (Sliders-FLUX) and TurboEdit tend to significantly alter the subjects' identities, as seen in the eyeglass and age transformations. Although LEDITS++ manages to execute reasonable smile and eyeglass edits, it inadvertently changes the subject's identity in age-related edits. These results highlight the ability of our method to perform disentangled edits and maintain the identity of the subject while applying targeted edits. More comparisons with other editing methods are provided in the supplementary material. 

\subsection{Real Image Editing}

\begin{figure*}
    \centering
    \includegraphics[width=\linewidth]{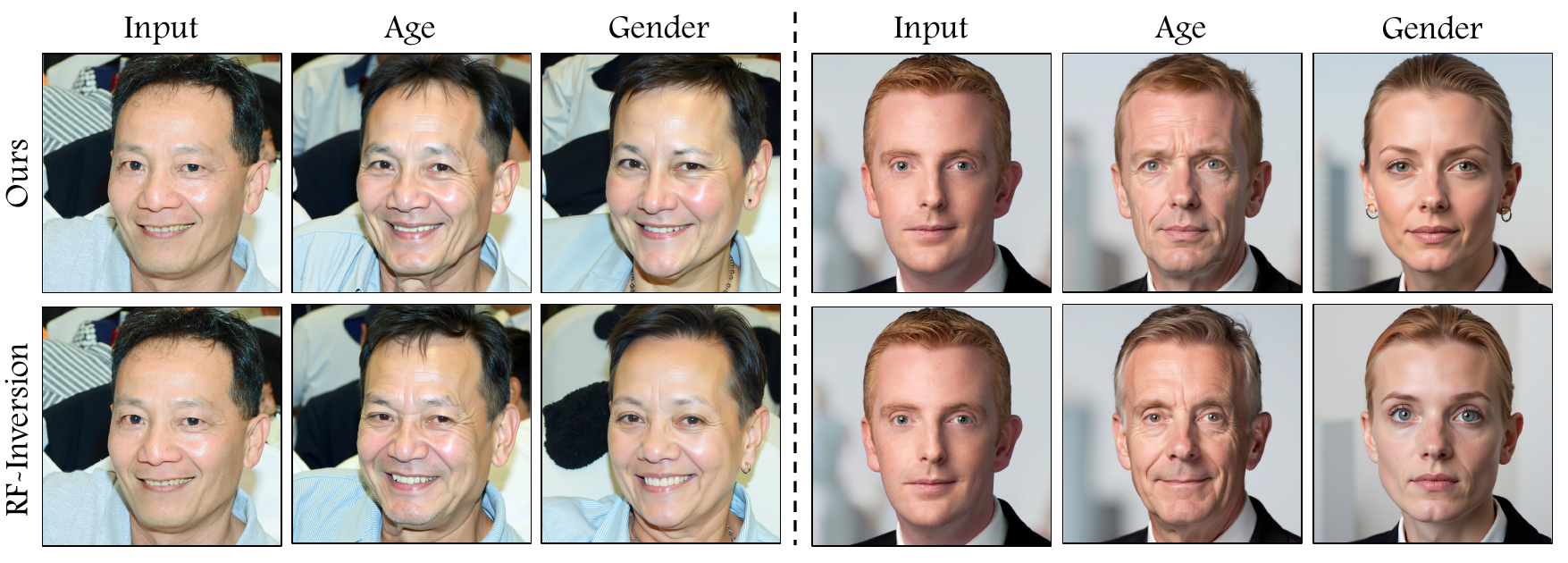}
    \caption{\textbf{Real Image Editing.} By integrating \textit{FluxSpace} on the inversion approach proposed by RF-Inversion \cite{rout2024rfinversion}, we extend our method for real image editing task. As we show qualitatively, our method achieves improved disentanglement over the performed edits compared to the baseline approach, where we use identical hyperparameters for the inversion task on both approaches.}
    \label{fig:real_images}
\end{figure*}

To be able to apply \textit{FluxSpace} to real images, we integrate our method into the inversion mechanism proposed by RF-Inversion \cite{rout2024rfinversion} and replace their editing with our editing algorithm. We provide real image editing results in Fig. \ref{fig:real_images}. However, we observe that, due to the nature of their inversion algorithm, the input image is not fully mapped into the latent distribution of the generator model (e.g. Flux), but rather consistency is enabled with a correction term. Since we focus on the interpretability of the latent space of rectified flow transformers, we focus on the models' output distribution and perform our experiments with images generated by Flux.

\subsection{Quantitative Results}
To further validate the effectiveness of our method, we performed quantitative experiments comparing it with state-of-the-art editing techniques, using Eyeglass and Smile semantics on a set of 60 images (see Table \ref{tab:quantitative}). We used metrics such as CLIP-T, CLIP-I \cite{radford2021learning}, and DINO \cite{caron2021emerging} for evaluation. CLIP-T assesses textual alignment by measuring the semantic similarity between the target description and the generated image, ensuring that the intended edits align with textual descriptors. CLIP-I and DINO evaluate image fidelity by measuring the visual similarity between the original and edited images, focusing on maintaining the identity and context of the original subjects.

The results indicate that our method excels in the CLIP-I and DINO metrics, showcasing its superior ability to retain the original identity while applying disentangled edits. Although methods such as TurboEdit are able to achieve higher CLIP-T scores, they do so at the cost of significant alterations to the subject's original identity. For instance, TurboEdit, despite its high CLIP-T score for eyeglasses, completely changes the individual's identity, as evidenced in Figure \ref{fig:qualitative}. This underscores our method's strength in balancing identity preservation with accurate semantic edits.

\noindent \textbf{User Study.} We conducted a user study involving 50 participants recruited through the Prolific.com crowd-sourcing platform. Each participant was presented with an original image alongside its edited version and the corresponding target text. They were then asked to assess how well the edited image captured the intended semantics while maintaining the original facial characteristics of the subject. This was rated on a Likert scale from 1 (strongly disagree) to 5 (strongly agree). The results in Table \ref{tab:quantitative} demonstrate that our method significantly outperforms competing approaches. More details about our user study are provided in the supplementary material.

\subsection{Ablation Studies}  

\begin{figure*}[t!]
    \centering
    \includegraphics[width=\linewidth]{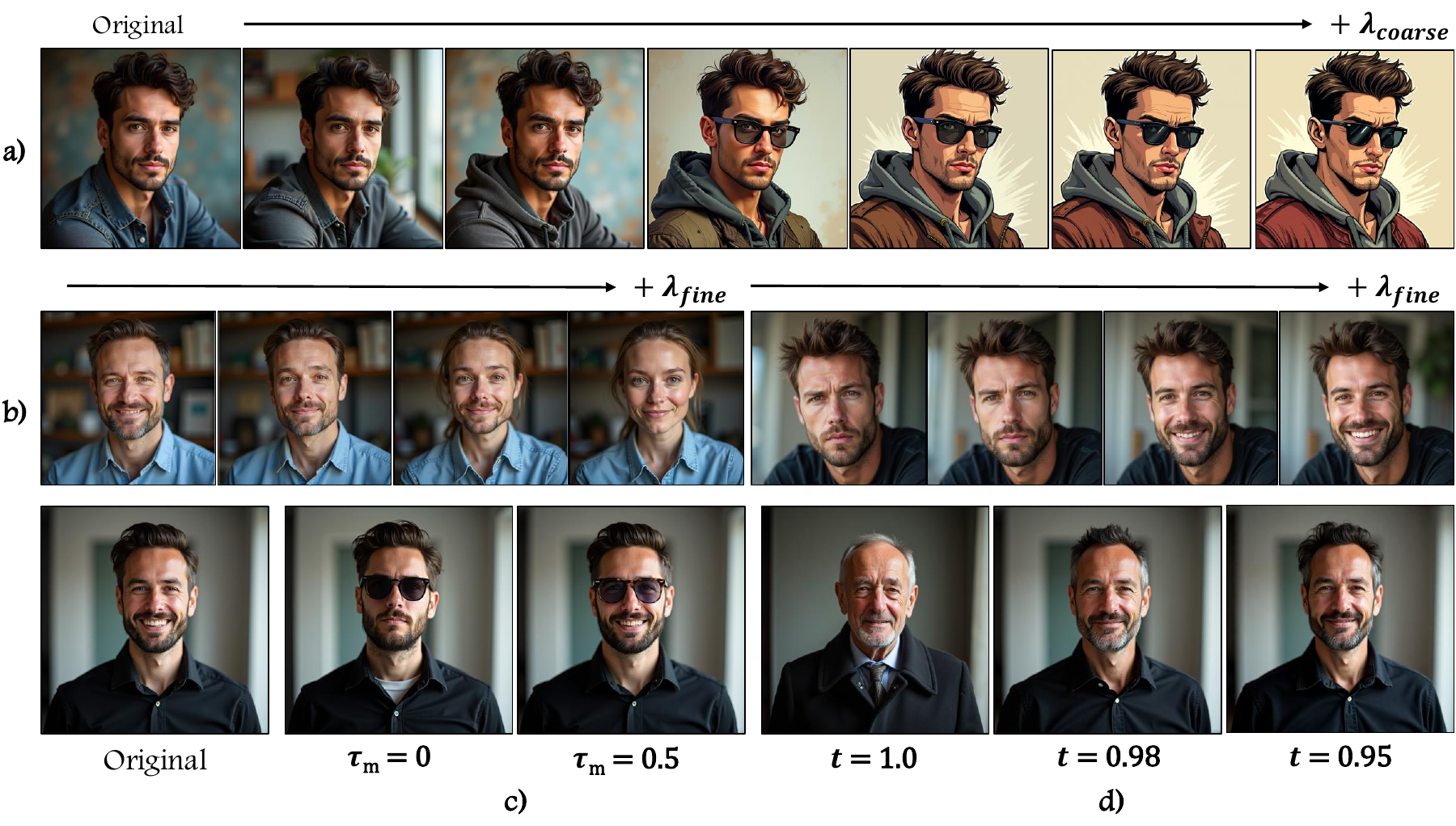}
    \caption{\textbf{Ablation Study.} We present ablations over the hyperparameters introduced within the FluxSpace framework. Specifically, we perform ablations on coarse editing scale $\lambda_{coarse}$, fine-grained editing scale $\lambda_{fine}$, masking coefficient $\tau_m$ and timestep $t$ when the editing is initiated. For all ablations, we report qualitative results for changing values of the specified hyperparameters.}
    \label{fig:ablations}
\end{figure*}

We perform ablation studies to demonstrate the effect of hyperparameters introduced by our method. Overall, we present ablations on fine-grained editing scale $\lambda_{fine}$, the coarse editing scale $\lambda_{coarse}$, the masking coefficient $\tau_m$, and the timesteps that the edit is applied. We provide qualitative results for each of these hyperparameters in Fig. \ref{fig:ablations}. \\

\noindent \textbf{Ablations on coarse editing scale.} We introduce the hyperparameter $\lambda_{coarse}$ to control the text embedding that determines the modulation parameters of the attention layer inputs \cite{esser2024scaling}. In our experiments, we identified that this representation is heavily involved in the overall appearance of the image. In Fig. \ref{fig:ablations}(a), we demonstrate the effect of $\lambda_{coarse}$ for controlling the pooled embedding for this modulation, where we use ``portrait photo of a man'' as the generation prompt, and ``comics style'' as the editing prompt. Upon interpolation within the scale [0.0, 1.0], we see that $\lambda_{coarse}$ enables controlling the style of the image, given the editing prompt. As we would like to identify the effect of $\lambda_{coarse}$ in isolation, we set $\lambda_{fine} = 0$ and $\tau_m = 0$ for all generations.

\noindent \textbf{Ablations on fine-grained editing scale.} We investigate the flexibility of content editing over the parameter $\lambda_{fine}$. Since the text embedding utilized in feature modulation only provides an approximate overview of the image to be generated, we associate $\lambda_{fine}$ with fine-grained editing, which determines the scale of the edit applied to the attention outputs. We provide data on the effect of $\lambda_{fine}$ in Fig. \ref{fig:ablations}(b), where we show semantic interpolation results on the attributes of gender and smile, with increasing values of the editing scale.

\noindent \textbf{Ablations on masking coefficient.} We introduce an optional attention masking mechanism to improve the disentanglement properties of the editing process. To demonstrate the effect of our masking strategy, we provide ablations in Fig. \ref{fig:ablations}(c), where the edit ``sunglasses'' is applied starting from the second denoising step, over a 30-step generation process. As can be observed from the presented results, our masking strategy improves the disentanglement of the edit and preserves the semantic properties of the original image better (e.g. facial expression).

\noindent \textbf{Ablations on Editing Timesteps.} Since our method allows the selection of editing timesteps, we perform an ablation over the effectiveness of the selection of timesteps where the edit is performed. As can also be observed from Fig. \ref{fig:ablations}(d), starting the editing process in earlier timesteps allows more drastic editing results, by sacrificing consistency with the original image. This is because the overall structure is still considered noisy from the perspective of the denoising model $\epsilon_\theta$. However, our method is still able to perform disentangled editing starting from early timesteps (e.g. skipping only 1-2 denoising steps).
\section{Limitation}
 
While FluxSpace offers significant benefits in controllable content generation and editing, it also introduces considerable ethical challenges. The ability of FluxSpace to perform precise image manipulation can lead to concerns over privacy, as individuals' likenesses might be used or altered without their explicit consent, a limitation that is commonly shared by other editing methods \cite{korshunov2018deepfakes}. Furthermore, the potential for creating misleading or deceptive content raises issues regarding the authenticity and trustworthiness of digital media.  These concerns are amplified by the capacity of FluxSpace to perform highly realistic and convincing edits that might be indistinguishable from the original images. This capability could be misused in scenarios such as creating fake news, impersonating individuals in harmful ways, or manipulating public opinion by altering perceptions of reality.  Given these risks, it is imperative to develop and implement robust ethical guidelines and regulatory frameworks that ensure the responsible use of image manipulation technologies.   Despite these risks, our approach highlights the internal knowledge of generative models for the task of image editing, which could facilitate research aimed at limiting harmful applications.
\section{Conclusion}

In conclusion, we propose FluxSpace, a novel method that demonstrates a robust ability to perform targeted, disentangled edits across a range of attributes and styles while maintaining the original identity of subjects in images. The qualitative and quantitative results, supplemented by a user study, underscore its effectiveness in preserving the characteristics of the edited image and achieving intended semantic changes, outperforming several state-of-the-art methods. Moreover, FluxSpace excels in performing fine-grained edits across various domains, along with stylistic transformations, showcasing its versatility and wide applicative potential in the field of image editing. 
\newpage
{
    \small
    \bibliographystyle{ieeenat_fullname}
    \bibliography{main}
}

\clearpage

\setcounter{section}{0}
\renewcommand{\thesection}{\Alph{section}}

\maketitlesupplementary

\section{Joint Transformer Block Architecture}
\label{sec:joint_block}
In our editing algorithm introduced in Sec. \ref{sec:methodology}, we focus on joint transformer blocks in Flux, where text and image inputs are processed in their respective representation spaces. As noted in \cite{esser2024scaling}, this design enables each modality to operate within its own feature space, while an interaction mechanism - specifically, the attention layer in the MM-DiT architecture - combines the two. Unlike single transformer blocks, which do not distinguish between feature spaces of different modalities, joint transformer blocks explicitly assume distinct spaces for text and image representations. Based on this difference, we conceptualize these blocks as the areas in which text content semantically influences image content. Consequently, we define our edits within these blocks, leveraging their unique capacity for cross-modal integration. Below, we outline the key components of the joint transformer block architecture in Flux to provide a better overview of our editing approach.

\paragraph{Modulation Mechanism.} The MM-DiT architecture, based on the DiT framework \cite{peebles2023scalable}, begins by performing a modulation operation utilizing coarse conditioning $c_{pool}$, timestep embedding $t_{emb}$ and guidance scale embedding $g_{emb}$. These embeddings are combined to compute the modulation embedding $m$, as described in Eq. \ref{eqn:modulation}, within the MM-DiT architecture used in Flux.

\begin{equation}
    \label{eqn:modulation}
    m = c_{pool} + t_{emb} + g_{emb}
\end{equation}

\paragraph{Attention Computation.} Given the modulation embedding $m$, input image features $x$, and contextual text features $c_{ctxt}$, the features undergo a modulation operation before the attention process. Subsequently, the text and image features are normalized using their respective layers. Using these normalized features, the joint attention block computes the attention for both modalities, using the query ($Q$), key ($K$), and value ($V$) features for the text and image inputs. Throughout this paper, we refer to the combination of modulation, normalization, and attention computation as a single pass through the attention layer, parameterized by $\theta$, denoted as $l_\theta(x, c, t)$ for timestep $t$.

\section{User Study Details}

\begin{figure*}
    \centering
    \includegraphics[width=0.9\linewidth]{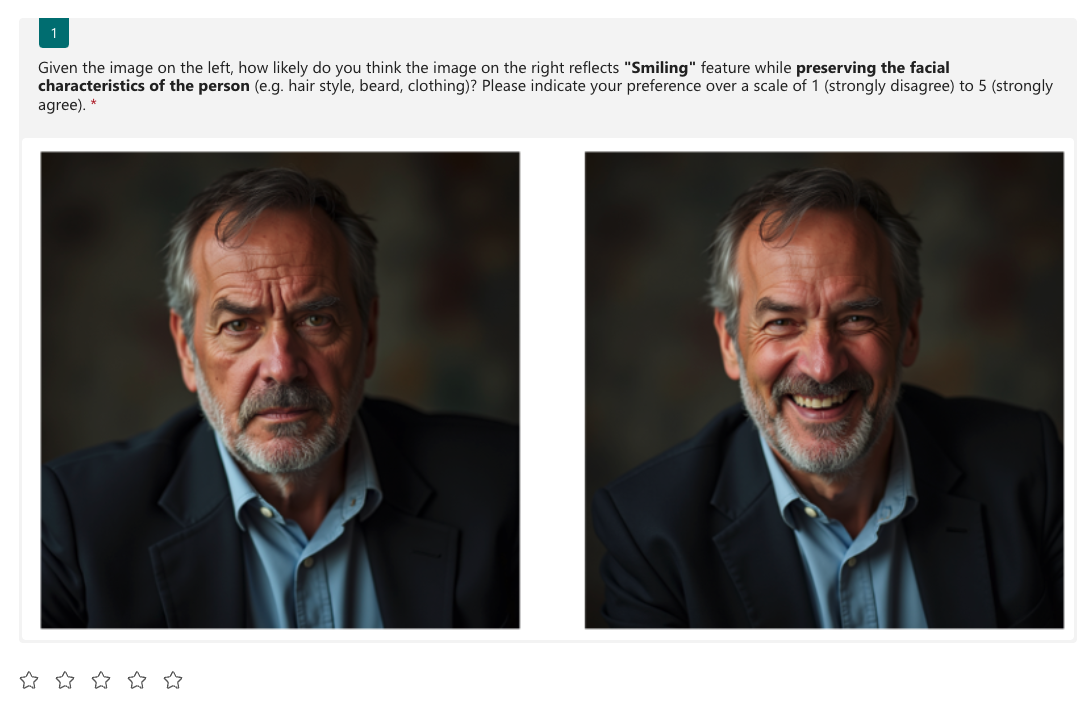}
    \caption{\textbf{User Study Setup.} We conduct our user study on unedited-edited image pairs. For each editing method, we provide the original image where the edit is not applied, with the edited image, and ask the users to rate the edit from a scale of 1-to-5. On the Likert scale that the users are asked to provide their preference on, 1 corresponds to unsatisfactory editing and 5 corresponds to a satisfactory edit.}
    \label{fig:user-study-setup}
\end{figure*}

To perceptually evaluate our method against competing approaches, we conducted a user study with 50 participants on the Prolific platform, where the results are provided in Table \ref{tab:quantitative}. For further clarification of the user study conducted, we provide a sample question in Fig. \ref{fig:user-study-setup}. In each question, we ask the users to rate the edit on a scale of 1-to-5 (1 for unsatisfactory, 5 for very satisfactory), after providing the image before editing and the edited image.

\section{Details on Quantitative Comparisons}

In this section, we provide the details for the quantitative results provided in Table \ref{tab:quantitative}. For clarity, we explain the setup used for each of the compared approaches, the models used for evaluation, and the hyperparameters used for editing with \textit{FluxSpace}.

\subsection{Competing Methods}

We compare the editing capabilities of \textit{FluxSpace} with RF-Inversion \cite{rout2024rfinversion}, Sliders-FLUX \cite{gandikota2023concept}, TurboEdit \cite{deutch2024turboedittextbasedimageediting} and LEDITS++ \cite{brack2024ledits++}. Below, we explain the setup used for each of these methods along with implementation details where necessary.

\paragraph{RF-Inversion \cite{rout2024rfinversion}.} Using the algorithm provided in \cite{rout2024rfinversion}, we re-implement RF-Inversion using \texttt{diffusers} library. For the editing hyperparameters, we use the parameter set provided by the authors for the eyeglasses edit. Even though the authors do not provide the exact hyperparameters for the smile edit in their paper, we utilize the hyperparameters used for the age editing task. Specifically, for the hyperparameters starting time ($s$), stopping time ($\tau$), and strength ($\eta$) we use the values \textbf{6, 25, 0.7} for the eyeglasses edit and \textbf{0, 5, 1.0} for the smile edit. For all generations, we use 30 steps, consistent with the setup used for \textit{FluxSpace}. Comparing the results presented in Table \ref{tab:quantitative} and the results provided in \cite{rout2024rfinversion}, our reported results are consistent with their quantitative evaluation, where we consider our implementation successful. As we also demonstrate qualitatively in Fig. \ref{fig:qualitative} and \ref{fig:real_images}, our approach succeeds over this baseline by improving the disentanglement and editing capabilities, where both approaches use \texttt{FLUX.1-dev} as the generative model.

\paragraph{Sliders-FLUX \cite{gandikota2023concept}.} Even though Concept Sliders was originally developed for UNet-based diffusion models \cite{podell2023sdxl, rombach2022high}, we utilize the extension of this method on rectified flow transformers. To do so, we use the \texttt{sliders} implementation provided for Flux, by the authors\footnote{\url{https://github.com/rohitgandikota/sliders/tree/main/flux-sliders}}. In all of our experiments, we use \textit{text} sliders which we train for eyeglasses and smile attributes. For training, we follow the default hyperparameters provided in the official implementation, where the LoRA rank is set to 16 and training is performed for 1000 iterations for all experiments, using \texttt{FLUX.1-dev}. Demonstrated in Fig. \ref{fig:qualitative} and Table \ref{tab:quantitative}, Sliders-FLUX struggles with the preservation of the input subject identity where significant alterations are observed during editing (see Fig. \ref{fig:qualitative}, middle row). We relate this issue with the training performed for the LoRA adapters and the fact that the edited concept cannot be clearly isolated. Note that our method does not require such training for the image editing task.

\paragraph{TurboEdit \cite{deutch2024turboedittextbasedimageediting}.} As a method based on few-step text-to-image generation models, we perform comparisons with TurboEdit \cite{deutch2024turboedittextbasedimageediting} which uses SDXL-Turbo \cite{sauer2025adversarial}. Following the official implementation of TurboEdit, we perform our comparisons with resolution 512 x 512 and 4 inference steps. Regarding the hyperparameters of the method, we use the pseudo-guidance scale $w$ as 1.5, and the random seed for edits as 2\footnote{\url{https://github.com/GiilDe/turbo-edit/blob/master/main.py}}. Demonstrated in results presented in Fig. \ref{fig:qualitative}, TurboEdit succeeds in performing the edit but fails in disentanglement, where the edits ``age'' and ``eyeglasses'' result in significant edit-irrelevant changes (e.g. skin color in age edit).

\paragraph{LEDITS++ \cite{brack2024ledits++}.} We also compare our method with LEDITS++ \cite{brack2024ledits++}, which is the state-of-the-art semantic editing method for text-to-image diffusion models. In our comparisons, we use the version of LEDITS++ that uses SDXL\footnote{\url{https://github.com/ml-research/ledits_pp/tree/main}}. For all of our experiments, we use the editing guidance scale and the editing threshold values \textbf{5.0} and \textbf{0.75}, as we found the guidance scale effective for eyeglasses and smile edit and we do not change the threshold value from the default setup. Despite performing well in identity preservation and semantic editing tasks, LEDITS++ results in artifacts in the edited images, which are not clearly identified in the quantitative evaluation, but acknowledged in the user study conducted.

\subsection{Hyperparameter Selection}
We present the quantitative results for our method in Table \ref{tab:quantitative}. In our experiments, we use a fixed set of hyperparameters for each edit evaluated, which are coarse editing scale $\lambda_{coarse}$, fine-grained editing scale $\lambda_{fine}$, mask threshold $\tau_m$, and starting iteration for edit $i$. The hyperparameter sets we used for the ``eyeglasses'' and ``smile'' edits are as follows:
\begin{itemize}
    \item \textbf{Eyeglasses:} $\lambda_{coarse} = 0.8$, $\lambda_{fine} = 5$, $\tau_m = 0.5$, $i = 3$ 
    \item \textbf{Smile:} $\lambda_{coarse} = 0.5$, $\lambda_{fine} = 8$, $\tau_m = 0.5$, $i = 5$ 
\end{itemize}

We use the editing prompts ``eyeglasses'' and ``smiling'' to perform these edits. Note that even though these edits require fine-grained changes in the image, our approach can apply these edits in a disentangled way even when the editing process starts in early timesteps. For all of our experiments, we set the number of inference steps as 30. We use a fixed seed of 0 in all of our experiments.

\subsection{Metrics Used}
As we also specify in Table \ref{tab:quantitative}, we use CLIP-T, CLIP-I, and DINO metrics to quantitatively evaluate our method. To enable the reproducibility of our experiments, we also share details on the model weights we use to obtain these scores. Specifically, we use \texttt{CLIP ViT-bigG/14}\footnote{\url{https://huggingface.co/laion/CLIP-ViT-bigG-14-laion2B-39B-b160k}} variant of the CLIP model to calculate the CLIP-I and CLIP-T scores. For DINO scores, we use \texttt{DINOv2}\footnote{\url{https://huggingface.co/facebook/dinov2-base}}.

\section{Supplementary Qualitative Results}

In this section, we provide supplementary qualitative results to further demonstrate the capabilities of our method.

\begin{figure*}
    \centering
    \includegraphics[width=\linewidth]{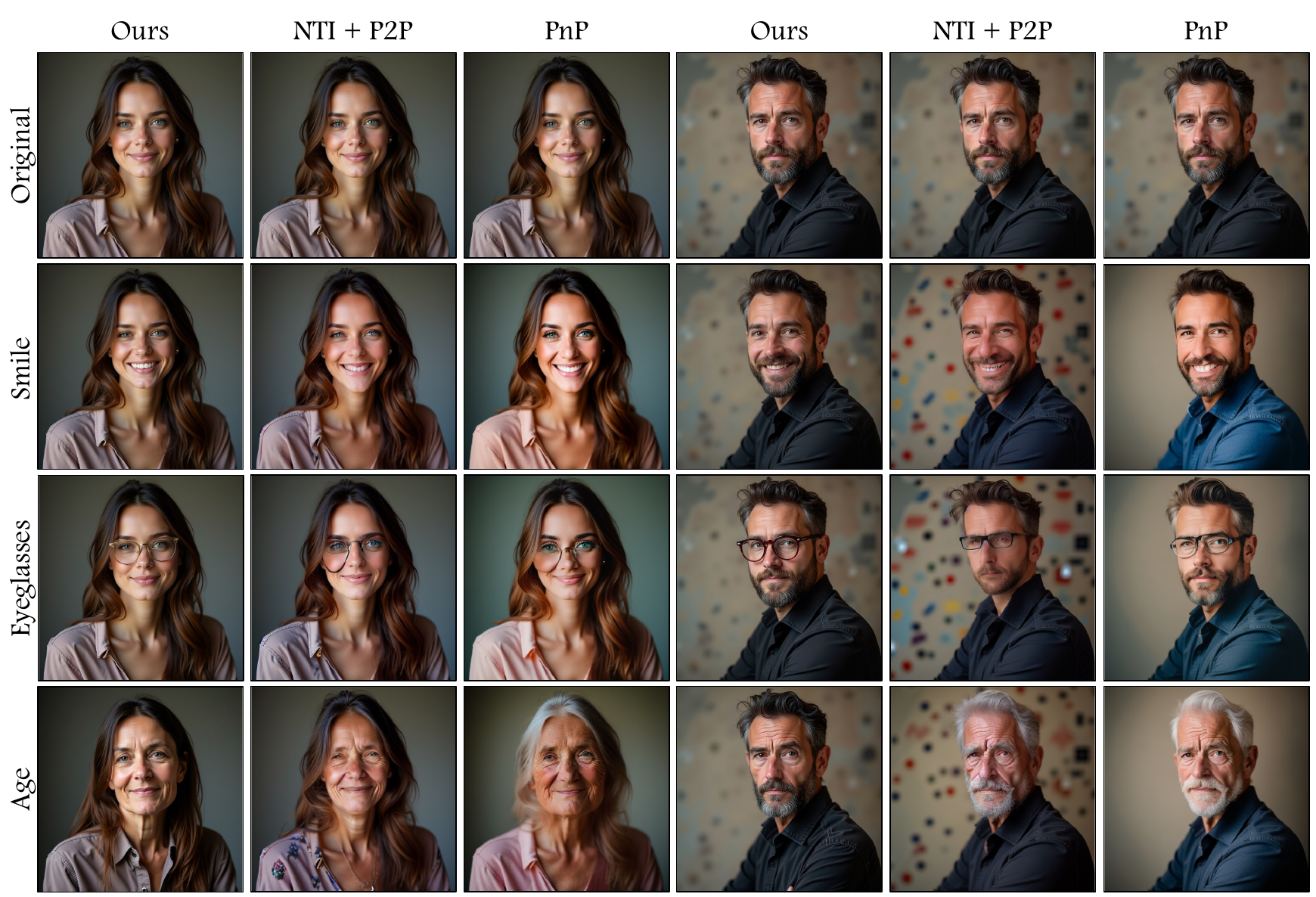}
    \caption{\textbf{Additional Qualitative Comparisons.} In addition to comparisons provided in the main paper, we provide additional comparisons with Prompt2Prompt \cite{hertz2022prompt} (with Null-Text Inversion \cite{mokady2022null}) and PnP-Diffusion \cite{tumanyan2023plug}, as Stable Diffusion based editing methods. As we demonstrate qualitatively, FluxSpace both achieves disentangled and semantically correct edits where competing methods contain artifacts in edited results (see the edit ``Eyeglasses'' for both methods), and significantly alter the subject identity (see ``Age'' edit).}
    \label{fig:supp_comparisons}
\end{figure*}

\paragraph{Supplementary Comparisons.} In addition to the comparisons provided in the main paper, we also compare the editing capabilities of \textit{FluxSpace} with methods based on Stable Diffusion. We compare our method with Prompt2Prompt \cite{hertz2022prompt} and PnP-Diffusion \cite{tumanyan2023plug} as competing approaches, where we perform inversion with Null-Text inversion \cite{mokady2022null} for Prompt2Prompt. We provide qualitative comparisons in Fig. \ref{fig:supp_comparisons}.

\begin{figure*}
    \centering
    \includegraphics[width=\linewidth]{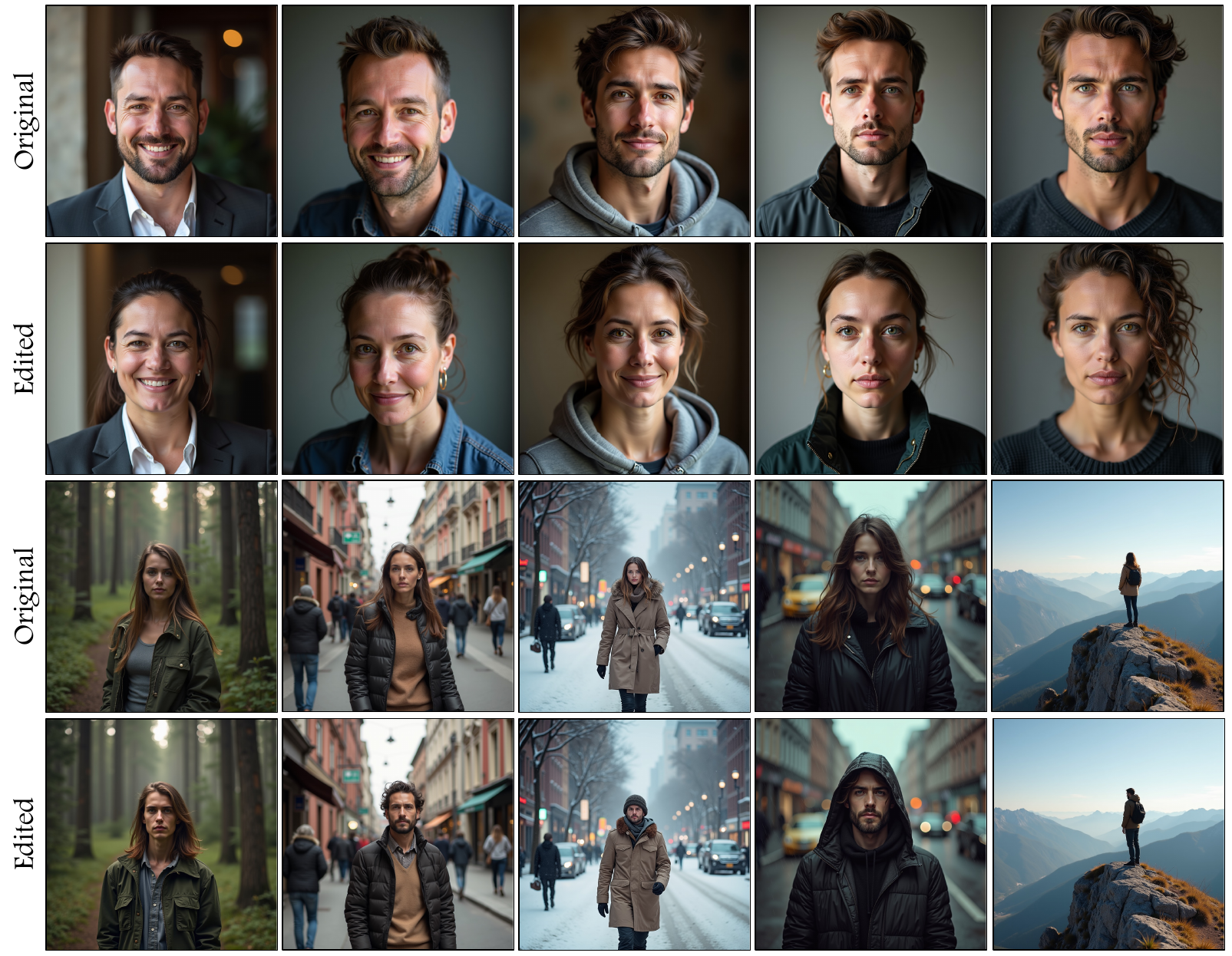}
    \caption{\textbf{Gender Editing Results.} We provide additional editing results for editing the gender semantics. As shown in the examples, our method succeeds in both male-to-female and female-to-male translations. We provide editing results on both portrait images, where our edits preserve the facial details, and edits on complex scenes where we succeed in only editing the human subject. Both in terms of preserving the identity of the subject and the background details, \textit{FluxSpace} succeeds in the disentanglement editing task.}
    \label{fig:supp_gender}
\end{figure*}

\begin{figure*}
    \centering
    \includegraphics[width=\linewidth]{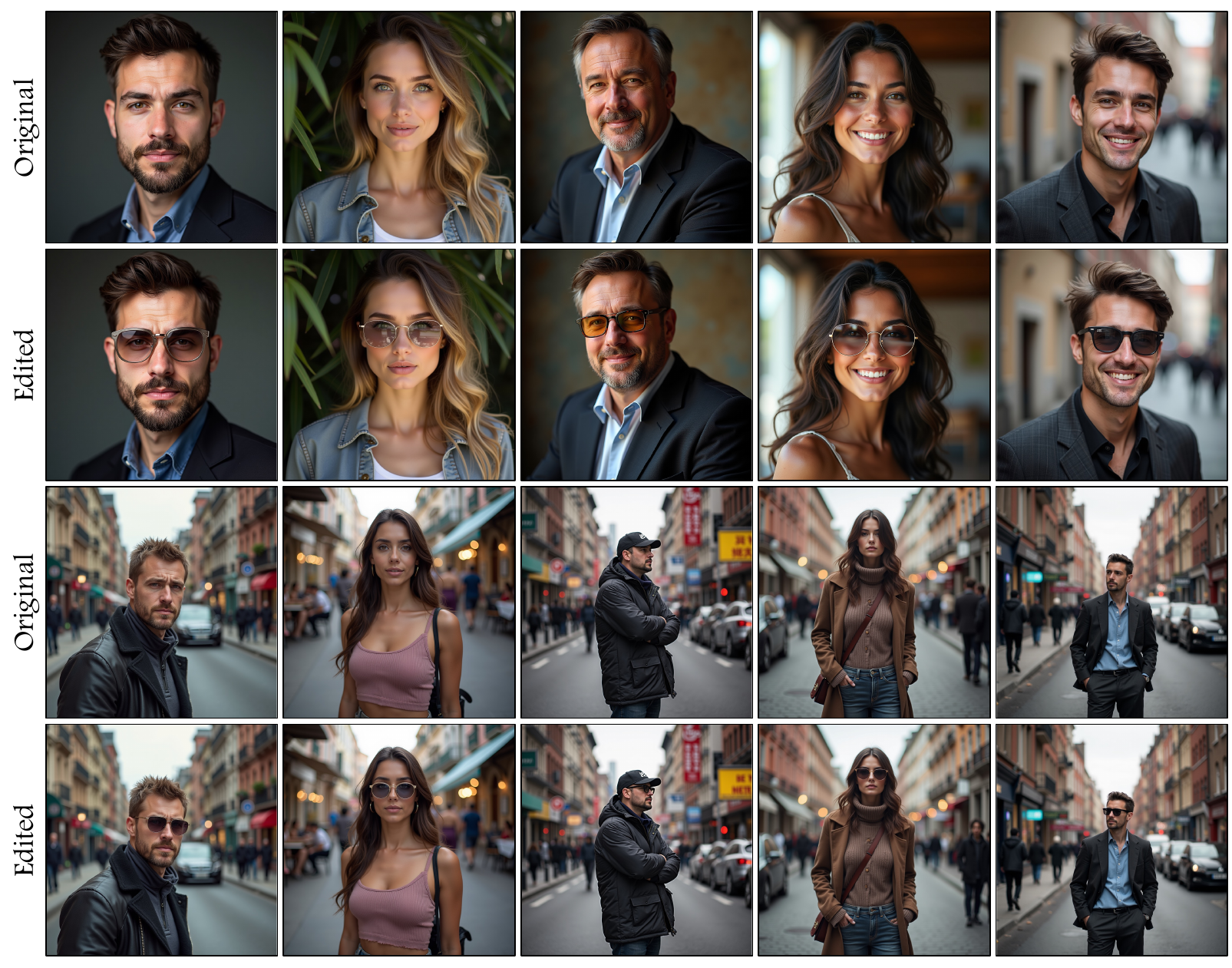}
\caption{\textbf{Sunglasses Editing Results.} We provide additional qualitative results for the edit ``adding sunglasses''. As we demonstrate on human subjects in both portrait images and more complex scenes, our editing method can accurately target where the edit should be applied without any input mask. We show the editing capabilities of \textit{FluxSpace} both in images where the human subject is the main focus of the image (first two rows) and with human subjects as a part of a scene (last two rows). In both cases, our method succeeds in performing the desired edit and preserving the edit-irrelevant details.}
    \label{fig:supp_sunglasses}
\end{figure*}

\begin{figure*}
    \centering
    \includegraphics[width=\linewidth]{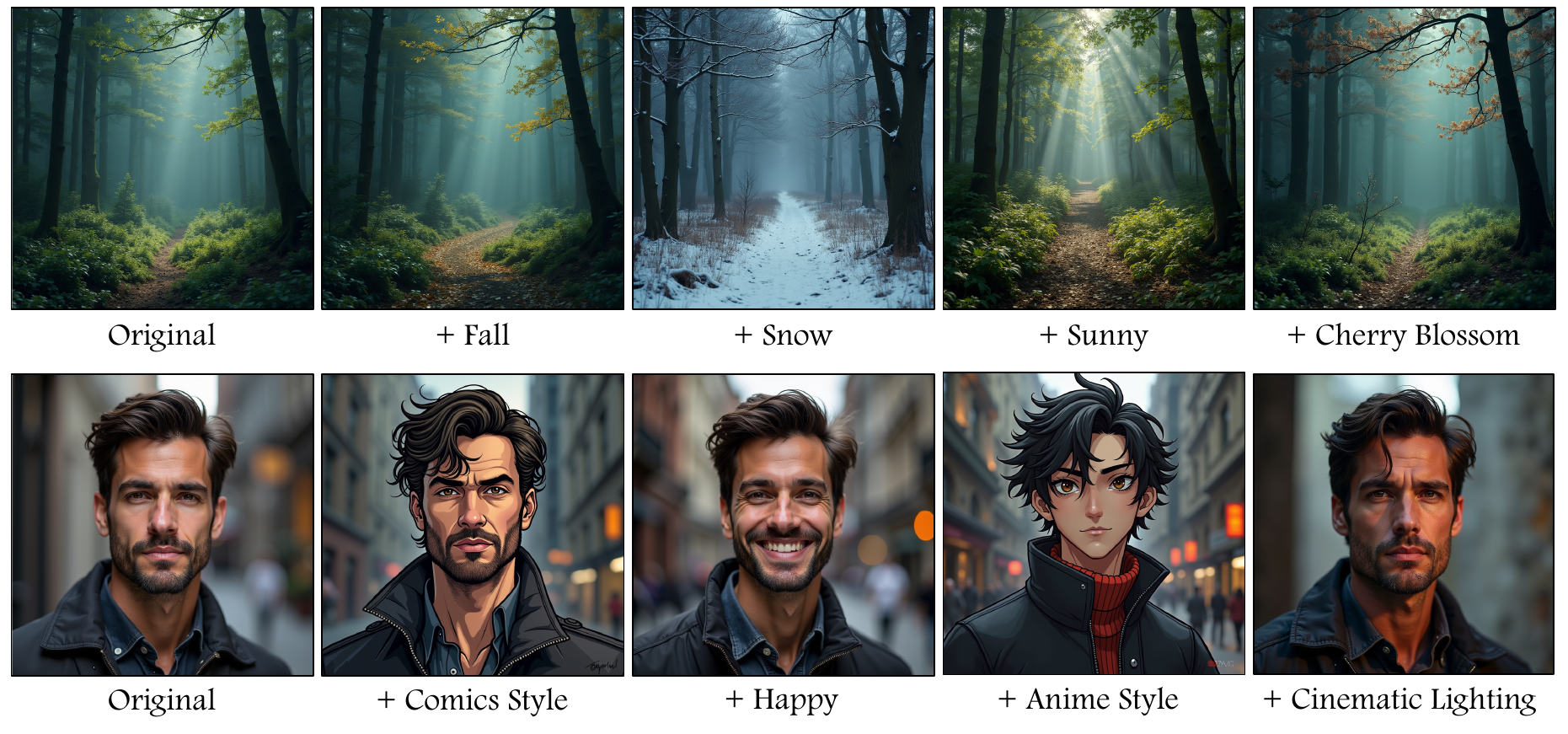}
    \caption{\textbf{Conceptual Editing Results.} We provide editing results with abstract concepts, that affect the overall appearance of the image. Our method succeeds in performing edits that alter the content of the image (top row) by being able to interpret the structures in the unedited image (e.g. the trees on the back for the edit ``cherry blossom'') and can change the style and overall appearance of the image (bottom row).}
    \label{fig:supp_concept}
\end{figure*}

\begin{figure*}
    \centering
    \includegraphics[width=\linewidth]{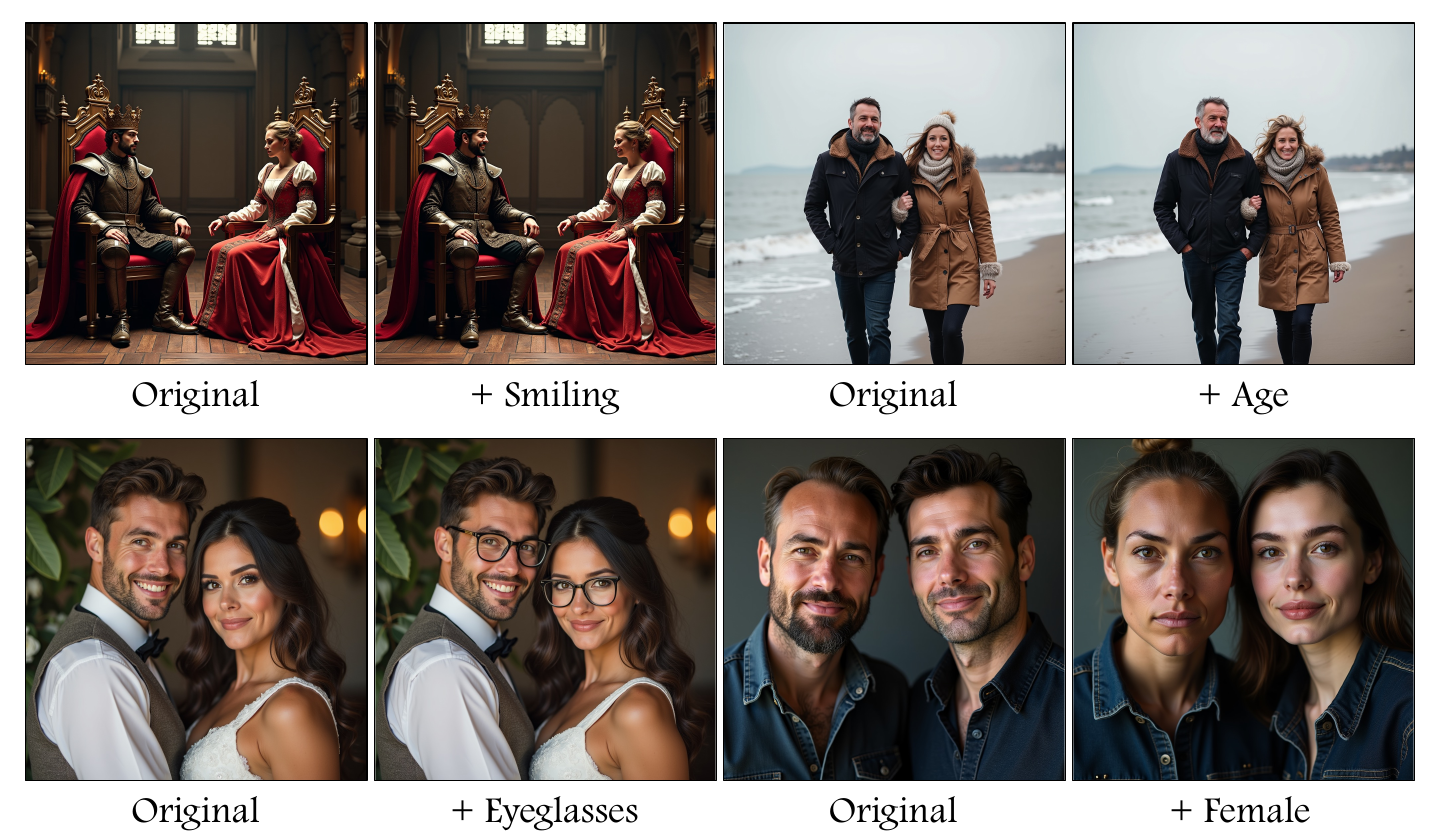}
    \caption{\textbf{Editing Results with Multiple Subjects.} We present qualitative results on images with multiple subjects. In addition to images with only one subject to be edited, \textit{FluxSpace} can apply edits by identifying semantics globally and editing multiple subjects at the same time. Note that our method does not use any external mask, and performs the edit completely with the semantic understanding of the rectified flow transformer.}
    \label{fig:supp_multi_subject}
\end{figure*}

\paragraph{Editing Examples.} Supplementary to the results provided in the main paper, we provide additional editing results in the supplementary material. We provide additional results for ``gender'' and ``sunglasses'' edits in Figs. \ref{fig:supp_gender} and \ref{fig:supp_sunglasses} for both portrait images and images in natural settings. Furthermore, we provide editing results for various concepts in Fig. \ref{fig:supp_concept} and with multiple subjects in Fig. \ref{fig:supp_multi_subject}.

\end{document}